\documentclass{article}

\usepackage[nonatbib, preprint]{neurips_2023}

\usepackage[utf8]{inputenc} 
\usepackage[T1]{fontenc}    
\usepackage{url}            
\usepackage{booktabs}       
\usepackage{amsfonts}       
\usepackage{nicefrac}       
\usepackage{microtype}      
\usepackage{xcolor}         

\usepackage{cite}
\usepackage{amsmath,amssymb,amsfonts}
\usepackage{mathtools}
\usepackage{graphicx}
\usepackage{textcomp}
\usepackage{xcolor}
\usepackage{numprint}
\npthousandsep{,}
\usepackage{fdsymbol} 

\usepackage[backref=page]{hyperref}
\usepackage{cleveref}
\usepackage{xstring}
\usepackage{xspace} 
\usepackage{csquotes}

\usepackage{adjustbox}
\usepackage{booktabs}
\usepackage{float}
\usepackage{subcaption}

\def\BibTeX{{\rm B\kern-.05em{\sc i\kern-.025em b}\kern-.08em
    T\kern-.1667em\lower.7ex\hbox{E}\kern-.125emX}}

\title{
    Comply: Learning Sentences with Complex Weights inspired by Fruit Fly Olfaction
}

\author{%
Alexei Figueroa$^{1}$ \quad Justus Westerhoff$^{1\diamondsuit}$ \quad Golzar Atefi$^{1\diamondsuit}$ \quad Dennis Fast$^1$ \\
\textbf{Benjamin Winter$^2$}\thanks{Work done at BHT.} \quad \textbf{Felix Alexander Gers}$^1$ \quad \textbf{Alexander Löser}$^1$ \quad \textbf{Wolfgang Nejdl}$^3$ \\
$^1$DATEXIS, Berliner Hochschule für Technik (BHT) \\
$^2$Medical AI Analytics \& Information GmbH \\
$^3$L3S, Leibniz University Hannover \\
\texttt{\{afigueroa,justus.westerhoff\}@bht-berlin.de} \\
$\diamondsuit$ Authors contributed equally.
}

\begin{document}

\newcommand{\methodname}{\texttt{Comply}\xspace}
\newcommand{\methodnametwo}{\texttt{ComplyM}\xspace}
\newcommand{\flyvec}{\texttt{FlyVec}\xspace}
\newcommand{\pflyvec}{\texttt{P-FlyVec}\xspace}
\newcommand{\biohash}{\texttt{BioHash}\xspace}
\newcommand{\transmlm}{\texttt{Transformer-MLM}\xspace}
\hyphenation{DDxGym}
\hyphenation{TwitterURLCorpus}
\hyphenation{SprintDuplicateQuestions}

\newcommand{\bert}{\texttt{BERT}\xspace}
\newcommand{\corpus}{Corpus\xspace}
\newcommand{\inputs}{Data\xspace}
\newcommand{\nvoc}{N_{\text{voc}}}
\newcommand{\pos}{\text{pos}}
\newcommand{\R}{\mathbb{R}}
\newcommand{\C}{\mathbb{C}}
\newcommand{\N}{\mathbb{N}}
\newcommand{\vect}[1]{\boldsymbol{\mathbf{#1}}}
\newcommand{\compconj}[1]{%
  \overline{#1}%
}

\maketitle
\begin{abstract}
    Biologically inspired neural networks offer alternative avenues to model data distributions. 
    FlyVec is a recent example that draws inspiration from the fruit fly’s olfactory circuit to tackle the task of learning word embeddings.
    Surprisingly, this model performs competitively even against deep learning approaches specifically designed to encode text, and it does so with the highest degree of computational efficiency.
    We pose the question of whether this performance can be improved further.
    For this, we introduce Comply.
    By incorporating positional information through complex weights, we enable a single-layer neural network to learn sequence representations.
    Our experiments show that Comply not only supersedes FlyVec but also performs on par with significantly larger state-of-the-art models.
    We achieve this without additional parameters.
    Comply yields sparse contextual representations of sentences that can be interpreted explicitly from the neuron weights.
\end{abstract}

\section{Introduction}\label{sec:introduction}
A large variety of deep learning models have centered on creating distributional language representations in recent years.
Among these, Transformer-based architectures such as \bert \cite{bert} have consistently set the state-of-the-art and are now considered baselines for Natural Language Processing (NLP) tasks.
However, better performance in metrics has come at the expense of a seemingly ever-growing computational complexity.
Naturally, there is a competing interest in developing efficient architectures, and often, these are inspired by intelligent biological systems that are efficient by nature.
Recently, \flyvec was proposed as a simple approach for creating contextual word embeddings motivated by the olfactory circuit of the fruit fly \cite{flyvec}.
These embeddings are binary vectors (hash codes) that are formed through neural inhibition via the $k$-Winner-Takes-All ($k$-WTA) activation.
While \flyvec is competitive with traditional word embedding methods like \texttt{word2vec} \cite{word2vec} and \texttt{GloVe} \cite{glove}, it is still superseded by contextual embedding methods such as \bert \cite{bert}.
Although \flyvec demonstrates an elegant approach that is more computationally efficient, it motivates the question of whether the performance can still be improved, while maintaining the same scale and architectural simplicity.

\par
We attempt to tackle this by, yet again, gathering inspiration from how fruit flies learn to identify odors.
Previous research has focused on the spatial bundling and modularity of neurons to create high-dimensional representations.
However, olfactory processing also involves processing temporal patterns, e.g., firing delays and duration. 
We highlight two ways in which fruit flies utilize temporal information.

\par
\textit{Relevance of odor arrival time}: 
The way fruit flies perceive a mixture of odors hints at how the arrival order of stimuli leads to better segregation and classification capabilities.
Multiple odors can be emitted either from the same source or from different sources.
Odors emitted from a single source have an approximately simultaneous arrival time due to the physical process of turbulent diffusion, as opposed to odors propagating from multiple sources, which results in distinct arrival times.
Several studies have shown that fruit flies \cite{10.1371/journal.pone.0036096, hopfield1991olfactory, SEHDEV2019113}, as well as other insects \cite{baker1998moth, saha2013spatiotemporal, nikonov2002peripheral, witzgall1991wind, andersson2011attraction}, likely use this asynchronous arrival time to perceptually segregate mixed odors.

\par
\textit{Using time as a medium to encode odors}:
Olfactory receptor neurons (ORNs) respond in a temporally heterogeneous manner that depends on both the type of odor and the type of ORNs \cite{raman2010temporally}.
This heterogeneity encompasses the firing rate and the time course (latency and duration), which results in odor identification irrespective of stimulus concentration \cite{stopfer2003intensity}.
Mutant flies that are ablated of spatial coding retain odor discriminatory capacity only through the temporal responses of ORNs \cite{dasgupta2008learned, olsen2007excitatory}.

\par
These two perspectives regarding time illustrate how spatial encoding (in the sense of neuron collocation) is \textit{not the only process} employed to generate odor embeddings.
We investigate the impact of integrating this temporal information in the quality of the embeddings.
Our method, \methodname, generalizes \flyvec to incorporate time information (word positions) in the analogy of mirroring text sequences (sentences) to odor segregation.
This is accomplished by expressing the input with phase differences and exploiting the semantics of complex numbers. 
Consequently, we learn a complex-valued parameter matrix through an extension of an unsupervised energy function.
We subject our models to a comprehensive evaluation in semantic sentence similarity tasks, as well as a text-based Reinforcement Learning environment.
Our experiments show how the features generated by \methodname boost the performance significantly, while preserving the parameter complexity, computational efficiency, and biological plausibility of \flyvec.
The main contributions of this work can be summarized as follows:
\begin{itemize}
    \item We introduce a novel architecture that can produce sparse encodings (binary hashes) incorporating positional information, while preserving computational efficiency and significantly boosting performance.
    \item Through a comprehensive quantitative evaluation of the unsupervised representations, we show the enhanced quality of the produced embeddings, being competitive with significantly larger architectures in NLP.
    \item We showcase qualitatively that the simplicity of our approach, \methodname, retains interpretable properties of \flyvec, while extending to the temporal dimension of sequences.
\end{itemize}
We make the source code\footnote{\href{https://github.com/DATEXIS/comply}{https://github.com/DATEXIS/comply}} available to reproduce the results of our experiments.
\section{Related Work}

\paragraph{Bio hashing.}
The mushroom body of the fruit fly is considered to be the center of olfactory learning and memory formation \cite{doi:10.1126/science.8303280}. 
It is where projection neurons (PN) make synaptic connections with Kenyon cells (KC).
In contrast to PNs, KCs have much more odor-selective responses \cite{turner2008olfactory, honegger2011cellular}.
In other words, only a small proportion of them becomes activated for a given odor due to inhibition \cite{lin2014sparse, papadopoulou2011normalization}.
Therefore, the KCs represent odors as sparse and distributed hash codes \cite{lin2014sparse}.
These codes are high-dimensional representations that the mushroom body output neurons (MBONs) use to form memories.
The sparsity and high specificity of these KC responses are thought to support the accuracy of this process\cite{laurent2002olfactory}. 

\par
The \flyvec\cite{flyvec} model repurposes bio hashing into learning word embeddings from raw text.
Using a one-layer neural network from PNs to KCs, it creates representations of a binarized input of context and target words.
The synaptic weights from PNs to KCs are learned with a biologically plausible unsupervised learning algorithm \cite{hopfield} inspired by Hebbian learning.
During inference, the $k$-Winner-Takes-All ($k$-WTA) mechanism creates sparse, binary hash codes.

\par
Our method, \methodname, uses a similar architecture and algorithm.
However, we augment the inputs with positional information and extend the loss and parameters accordingly.
Specifically, \methodname generalizes \flyvec, enabling bio hashes to represent sequences.

\paragraph{Text embeddings.}
The main goal of embedding text is to produce vectors that have a semantic correspondence to characters, words, or arbitrary sequences of text elements (tokens).
Examples are \texttt{word2vec} \cite{word2vec}, \texttt{GloVe} (Global Vectors) \cite{glove}, and \bert (Bidirectional Encoder Representations from Transformers) \cite{bert}.
The former two are examples of static word embeddings, which are constant regardless of context.
In contrast, \bert learns a function that can yield representations conditioned on the context: a contextual word embedding.

Transformer-based encoder models like \bert learn language representations using self-attention and masked language modeling among other objectives, which are not necessarily motivated by biological mechanisms.
These representations are \enquote{contextual}, i.e., they capture the ambiguity of words (sentences) given the context in which they appear.

The \flyvec model highlights how a biologically inspired architecture is also capable of learning contextual \textit{word} embeddings (binary hash codes).
In our work, we extend this model to \textit{sentence} representations.
We do this by learning positions as a complementary distributional property of text. 

\paragraph{Positional encoding.}
Recurrent Neural Networks (RNNs) \cite{rnn}, e.g., Long Short-Term Memory (LSTM) \cite{lstm, gers}, inherently capture positional information given that their state (and output) is always conditioned on the sequential input.
Similarly, Convolutional Neural Networks (CNNs) can implicitly learn to incorporate positional information \cite{cnn_position}.
Although self-attention \cite{attention} is designed to focus on different parts of the sequence depending on the task, it does not infer positional information.
When introducing Transformers, the work \cite{vaswani} explicitly incorporates positional encodings in an additive manner.
In fact, this component is crucial for the success of these architectures \cite{pos-information-crucial}.
Closer to our work, \texttt{RoPe} \cite{roformer} expresses positional information in Transformer models through a multiplicative approach.
Namely, an embedding vector of a token is \textit{rotated} based on its position.
Although this approach and ours are both non-parametric and leverage the complex plane to implement such rotations, we differ by explicitly expressing the input in the complex field and learning a complex parameter matrix.

\paragraph{Complex weights.}
Complex-valued Neural Networks (CVNNs) offer advantages in specific domains, as they can handle both amplitudes and phases of the signals.
Examples of such areas are image processing, and more generally signal processing where a datum is inherently complex.
For a comprehensive overview see, e.g., \cite{cvnn-survey}.
Close to our work, the method in \cite{complex_embeddings} uses a learned complex-valued map to enrich text embeddings with positional information.
In our work, using the semantics of the complex field yields a simple mathematical extension of the \flyvec model.
While CVNNs focus on end-to-end functions in the complex field, we use complex numbers only as a way of efficiently representing the order in sequences learned in neurons.
However, we do \textit{not} use complex activation functions, our embeddings are purely real and are constructed with the $k$-WTA mechanism.
\section{Methods}\label{sec:method}
\subsection{\flyvec\cite{flyvec}.}
\begin{figure}[tb]
    \centering
    \includegraphics[width=6.5cm]{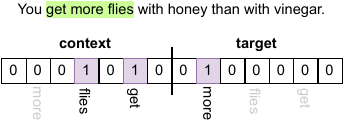}
    \caption{Sample input vector $\vect{v}\in\R^{2\nvoc}$ to the \flyvec model. The two blocks (context and target) of length $\nvoc$ are separated by the long vertical line in the middle. Here, \enquote{more} is the \textit{target word} and sliding window $\omega=3$. Adapted from \cite{flyvec}, Figure 2, p.3.}
    \label{fig:bag-of-words}
\end{figure}

This model adapts a single parameter matrix $\vect{W} \in \R^{K\times2\nvoc}$, modeling the PNs$\to$KCs synapses (projection neurons to Kenyon cells) and using a biologically inspired mechanism.
Here, $K$ is the number KCs, and $\nvoc$ is the size of the word vocabulary.
For training, every input sentence is split into sliding text windows of fixed length $\omega\in\N$ disregarding positions, thus modeling sets (bag-of-words).
The inputs $\vect{v} \in \R^{2\nvoc}$ are crafted as follows: the first $\nvoc$ components are the sum of one-hot encoded vectors of context tokens surrounding a \textit{target word}, and the second $\nvoc$ entries are a single one-hot vector corresponding to the \textit{target word}, see \cref{fig:bag-of-words}.
The parameter matrix $\vect{W}$ is learned by minimizing the energy function
\begin{equation}
E \coloneqq - \sum_{\vect{v} \in \inputs}
                       {
                         \frac
                              { 
                                  \langle
                                  \vect{W}_{\widehat{\mu}} , \vect{v}/\vect{p}
                                  \rangle 
                              }
                              { \langle
                                  \vect{W}_{\widehat{\mu}},\vect{W}_{\widehat{\mu}}
                                \rangle^{1/2} 
                              }
                       },
\label{eq:E}
\end{equation}
where
\begin{equation}
\widehat\mu \coloneqq \underset{\mu\in\{0,\dotsc,K-1\}}{\operatorname{argmax}} \ \langle \vect{W}_{\mu},\vect{v} \rangle,
\label{eq:argmax}
\end{equation}
$\vect{v}/\vect{p}$ is the input vector as described above divided by the word frequencies $\vect{p}\in\N^{\nvoc}$ in the whole training corpus, and $\inputs$ is the set of preprocessed inputs. 
The model emphasizes learning through lateral inhibition, i.e., during training, there is only a single neuron $\widehat\mu$ (a row $\vect{W}_{\widehat\mu}$ in the $\vect{W}$ matrix), which is adapted for a single sample.

\par
\subsection{\methodname}
Our extension revolvese around capturing the sequential nature of text data through added positional information.
We leverage the semantics of complex numbers and keep the same biologically inspired mechanisms, namely sparsity through $k$-Winner-Takes-All ($k$-WTA) activations.
We aim to preserve the properties of \flyvec regarding efficiency, simplicity, and interpretability.
Thus, we constrain ourselves to a single-parameter matrix.
Additionally, we focus on an extension that considers a complex-valued input, a single complex parameter matrix $\vect{W}$, and a compatible energy function $E$.

\begin{figure}[htb]
    \centering
    \includegraphics[width=7.95cm]{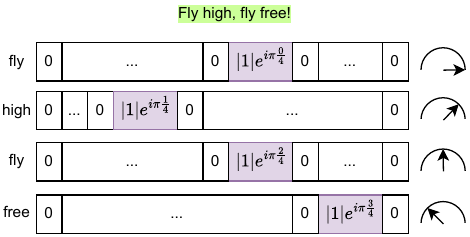}
    \caption{Sample input $\vect{z}\in\C^{L\times\nvoc}$ to our \methodname model, where $L=4$ is the length of the sample sentence \enquote{\textit{Fly high, fly free!}}.
    Positions are encoded as the complex $2L$th roots of one in the upper half-plane.}
    \label{fig:complex-sequence}
\end{figure}

\paragraph{Complex sequential input.}\label{ss:complexinput}
For each \textit{sentence} expressed as a concatenation of one-hot encoded vectors, we multiply each vector by $e^{i\pi\frac{l}{L}}$, where $L$ is the length of the sentence, $l \in \{0,1,\dotsc,L-1\}$ denotes the position of the respective word in the sentence and $i$ is the imaginary unit, see \cref{fig:complex-sequence}.
We denote this preprocessed input by $\vect{z} \in \C^{L\times\nvoc}$ and its complex conjugate by $\compconj{\vect{z}}$.
Note that this modification, compared to \flyvec, expands the dimension of the input by $L$, and reduces the length of each word vector by half since we do not have a target word as in \cref{fig:bag-of-words}.
Consequently, our parameter matrix $\vect{W}$ is in $\C^{K \times \nvoc}$ and has the same number of parameters as \flyvec ($2K\nvoc$), since for each complex parameter two real numbers are required.

With this modified input, an immediate extension of \cref{eq:E} with the Hermitian inner product $\langle\cdot,\cdot\rangle_H$ is not directly applicable, since for a sentence the quantity
\begin{equation}\label{eq:newquantity}
\sum_{l \in L}\langle \vect{W}_{\mu},\vect{z}_l \rangle_H = \sum_{l\in L}\vect{W}_{\mu}^T \compconj{\vect{z}}_l \ \in \C
\end{equation}
is complex and $\C$ is unordered, hence finding the $\operatorname{argmax}$ as in \cref{eq:argmax} is not possible.
Thus, we adapt $E$ to find a coherent maximally activated neuron $\widehat\mu$.

\paragraph{Adapting $E$.}
Due to the complex-valued nature of the Hermitian inner product, an alternative extension must be considered.
If we disregard word repetition, we can think of \cref{eq:newquantity} as $\langle \vect{W}_{\mu}, \tilde{\vect{z}}\rangle_H$, where $\tilde{\vect{z}} \coloneqq \sum_{l \in L}\vect{z}_l$ is a multi-hot encoded version of the input (similar to \flyvec).
We can decompose $\langle W_{\mu}, \tilde{\vect{z}}\rangle_H$ into two steps, element-wise multiplication and aggregation, and yield real values that are proportional to neuron activation and can be sorted:
\begin{itemize}
    \item \textit{Element-wise multiplication}: For every word and neuron, we perform a complex multiplication, which translates to scaling absolutes and subtracting phases. 
    The former indicates the correlation between the input \textit{intensity} and the neuron's output, while the latter gives us a measure of the distance (\textit{temporal difference}) between the word in neuron $\vect{W}_\mu$ and the sentence.
    \item \textit{Aggregation}:
    Simply adding these complex numbers does not retain positional difference information, since $e^{i\theta} = e^{i2\pi\theta}$ for any $\theta\in\R$.
    Hence, we instead sum the magnitudes and absolute values of phase differences to capture both the effect of maximum word activation (as in the \flyvec model) and the word position distance.
\end{itemize}
\Cref{eq:E-complex} reflects these modifications to the energy function $E$.
Note that the energy of the original \flyvec model (see \cref{eq:E}) is a special case of our approach when $\vect{W}$ is real and the inputs are preprocessed as in \cref{fig:bag-of-words}.
We slightly abuse notation by writing $L$ and $\{1,\dotsc,L\}$ interchangeably, and define
\begin{equation}
\begin{aligned}
E \coloneqq - \sum_{\vect{z} \in \inputs}\biggl(&\frac{ 
                               \sum_{l \in L}{
                                   \left\lvert \left\langle\vect{W}_{\widehat{\mu}}, \vect{z}_{l}/\vect{p}_{s_l}\right\rangle_H \right\rvert
                                   }
                             }
                             {  
                                \left\langle
                                \vect{W}_{\widehat{\mu}},\vect{W}_{\widehat{\mu}}
                                \right\rangle^{1/2}_H 
                             } \\
                            &+ 
                            \sum_{l \in L}{
                             \left\lvert \operatorname{Arg}\left\langle\vect{W}_{\widehat{\mu}}, \vect{z}_l/\vect{p}_{s_l} \right\rangle_H\right\rvert
                             } 
                        \biggr),
\label{eq:E-complex}
\end{aligned}
\end{equation}
where
\begin{equation}
\widehat\mu \coloneqq \underset{\mu\in\{0,\dotsc,K-1\}}{\operatorname{argmax}} \sum_{l \in L}{\left\lvert \left\langle\vect{W}_{\mu}, \vect{z}_{l} \right\rangle_H \right\rvert + \left\lvert \operatorname{Arg}\left\langle\vect{W}_{\mu}, \vect{z}_l\right\rangle_H\right\rvert, }
\label{eq:mu-complex}
\end{equation}
$\operatorname{Arg}$ is the function that extracts the phases from the complex numbers, and $s_l$ is the index in the vocabulary of the $l$th word in a sentence $[s_0,...,s_{L-1}]$.
\Cref{eq:E-complex} favors learning complex weights for words in one half of the complex plane, since the absolute angle difference $\lvert \operatorname{Arg}\cdot\rvert$ is maximized, and its maximum is at $\pi$.
If we replaced the additions in both \cref{eq:E-complex,eq:mu-complex} with subtractions, the synapses would be learned on the same half-plane as the input vectors $\vect{z}$.
Either option is of no consequence for the model and we choose the additive expression for our experiments.
Furthermore, we highlight that the two summands in \cref{eq:mu-complex} are not of the same scale; the former, $\lvert \langle\vect{W}_{\mu}, \vect{z}_{l} \rangle_H\rvert$, is in $[0, \infty)$, whereas the latter, $\lvert \operatorname{Arg}\langle\vect{W}_{\mu}, \vect{z}_l\rangle_H\rvert$, is in $[0, \pi]$.
To consider this, we experimented with replacing addition ($+$) by multiplication ($\cdot$) in \cref{eq:E-complex} and \cref{eq:mu-complex}. 
This however resulted in instability leading to model overfitting (multiple neurons learn the same sequences) and, more prominently, would not constitute a generalization of the original energy presented in \cref{eq:E}.
Nevertheless, in the case of inference, we also test the multiplicative approach. 

\par
Note that, as depicted in \cref{fig:complex-sequence}, we limit the phases in the input $\vect{z}$ to lie in $[0,\pi)$ to enforce learning phases in $\vect{W}$ that also only span half of the complex plane.
This resolves the ambiguity of having distinct positions in word phases that would be mapped equivalently in the phase distance component of our energy function, e.g., a word appearing in $\frac{-\pi}{2}$ and $\frac{\pi}{2}$ and the weight of that word in $\vect{W}_{\mu}$ with phase $0$.

\par
To summarize, the semantics and operations of complex numbers naturally give us a framework to generalize the objective of minimizing $E$, while capturing positional information.

\begin{figure*}[tb]
    \centering
    \includegraphics[width=1.0\linewidth]{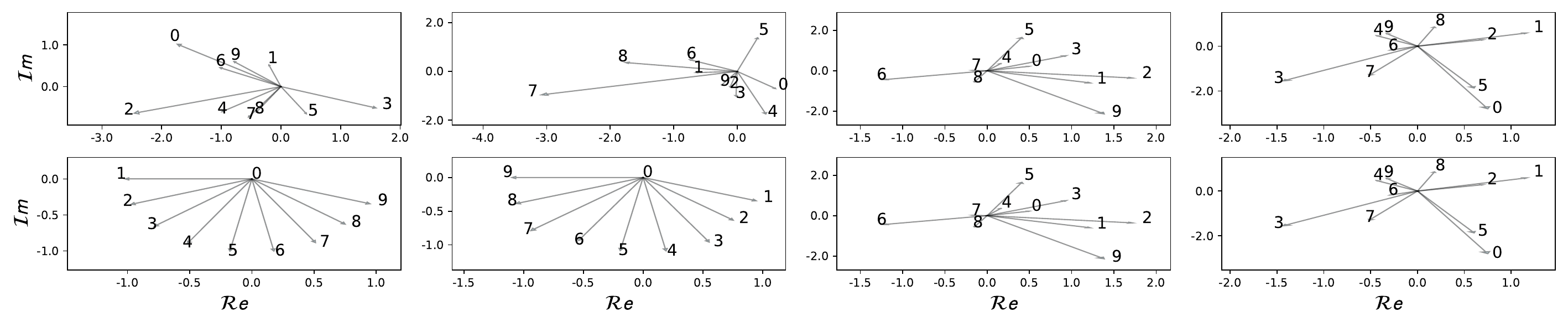}
    \caption{Toy example of training with two sequences $1,2,\dotsc,9$ and $9,8,\dotsc,1$ modified as in \cref{fig:complex-sequence} with phases in $[0, \pi)$.
    Top row: The randomly initialized complex weights of a model with four Kenyon cells.
    Bottom row: Weights after learning with \cref{eq:E-complex}. The loss makes the model memorize the sequences in the lower half of the complex plane (phases in $[-\pi,0)$) in the first two neurons.
    The other neurons remain unchanged.}
    \label{fig:toy}
\end{figure*}

\paragraph{Toy example.}\label{ss:toy}
We illustrate the mechanics of the loss presented in \cref{eq:E-complex} with a toy example that uses only four KCs ($K=4$) and ten words (digits) ($\nvoc=10$) for two sentences (sequences): $1,2,\dotsc,9$ and $9,8,\dotsc,1$. The model is trained with backpropagation following \cref{eq:E-complex}.
\Cref{fig:toy} shows the initial state in the complex plane for each neuron, as well as the converging state after learning.
Both sequences are explicitly learned in the network in the opposite half-plane of the input (that is, with angles in $[-\pi;0)$). 
The $1$-WTA activation (\cref{eq:E-complex}) successfully creates a sparse representation of the two sequences \textit{without interference}. 
This can be observed since the remaining neurons are unchanged and the two sequences are learned (\textit{imprinted}) in two distinct neurons.

\paragraph{Binary hashes.}
\label{ss:hashes}
We construct binary feature vectors for \methodname from the order of activations of the KCs.
Specifically, as in \cite{biohash}, the output vector $\vect{h} \in \{0,1\}^K$ of an input $\vect{z}$ is defined for all $\mu \in\{1,\dotsc,K\}$ as
\begin{equation}
    \vect{h}_\mu \coloneqq
    \begin{cases}
        1, & \text{if } \sum_{l \in L}{\left\lvert \left\langle\vect{W}_{\mu}, \vect{z}_{l} \right\rangle_H \right\rvert + \left\lvert \operatorname{Arg}\left\langle\vect{W}_{\mu}, \vect{z}_l\right\rangle_H\right\rvert} \\ & \hphantom{if }\text{is in the top-$k$ of all KC activations,}\\
        0 & \text{otherwise,}
    \end{cases}
    \label{eq:hash}
\end{equation}
where $k \in \N$, the hash length, is a hyperparameter.
Since the two summands are, as noted earlier, of different scales, we also experiment with a version we call \methodnametwo that does multiplication ($\cdot$) instead of addition ($+$) in \cref{eq:hash} above.
\section{Experiments}\label{sec:experiments}
We evaluate our methods in sentence semantic similarity tasks, as well as in a text-based Reinforcement Learning (RL) environment.

\subsection{Baselines}
We focus on \flyvec since it is the architecture we extend.
For the sentence similarity tasks, we also compare against \bert embeddings which are contextual representations based on the Transformer architecture.
We use the \texttt{bert-base-uncased} weights publicly available on \texttt{Hugging Face}.\footnote{\url{https://huggingface.co/google-bert/bert-base-uncased}}
Likewise, we use the publicly available weights of \flyvec. \footnote{\url{https://pypi.org/project/flyvec/}}
Following the evaluation in \cite{flyvec}, we constrain the tokenizer of \bert to the top \numprint{20000} word indices for a fair comparison against \flyvec and our models.

\par
In the RL evaluation, we compare against a \flyvec model that we pretrain on PubMed (\pflyvec), and the best performing Transformer-based encoder model in \cite{ddxgym}, denoted as \transmlm.
This architecture provides a policy with an auxiliary language modeling objective that is crucial to stabilize Transformer-encoder learning in online RL settings.
We use this approach for both \pflyvec and \methodname.

\subsection{Data and model training}\label{ss:pretraining}
We use large text corpora to train the models for our experiments.
Sentences are the largest sequence structure that we model with \methodname (see \cref{ss:complexinput}) and we create batches accordingly.
Although the sliding window approach is only relevant for \flyvec (see \cref{fig:bag-of-words}), we use it in our implementation only to construct batches of the same shape (phases of the input are independent of this), and guarantee that \methodname sees tokens as often as \flyvec. 
We pretrain our models on two different text corpora depending on the domain of the target task:

\paragraph{Open Web Corpus (OWC) \cite{openweb}.}
For the sake of comparability with \cite{flyvec}, we use the OWC dataset to train \methodname.
It consists of roughly eight million web documents filtered to be longer than 128 tokens in English, resulting in about 40GB of text files.
We preprocess and encode these text data with the \flyvec tokenizer provided in \cite{flyvec}.
It maps every word to a single integer in a vocabulary of \numprint{20000} IDs, resulting in approximately 6.5B tokens.
This model is evaluated against the semantic similarity tasks.

\paragraph{PubMed.}
We construct a dataset from the PubMed database\footnote{\url{https://pubmed.ncbi.nlm.nih.gov/download/}} keeping only abstracts of articles as in \cite{pubmedbert}.
Their model, \texttt{PubMedBERT}, highlights that in-domain abstracts are sufficient to achieve competitive results in biomedical tasks.
We filter for abstracts in English and use this tokenizer to yield approximately 6.9B tokens.
We pretrain both \pflyvec and \methodname with these data.
The resulting models have the same number of parameters and we use them in the RL evaluation. 

\subsection{Semantic similarity evaluation}
We emphasize that \methodname creates representations for sentences.
We are interested in evaluating the pretrained embeddings learned in an unsupervised manner \textit{without further fine-tuning}.
Hence, we select the subset of tasks compiled in the Massive Text Embedding Benchmark (MTEB) \cite{mteb} that targets sentence embeddings without additional models (classification heads or clustering algorithms).
Furthermore, we focus on tasks comprising English sentences. 
This results in the majority of the Semantic Textual Similarity (STS), as well as Pair Classification tasks.
\par
The STS tasks are the SemEval workshop tasks STS12-17 in addition to BIOSESS, SICK-R, and STSBenchmark.
In these tasks, the similarity between two sentences is assessed based on the embeddings produced by a model.
For \flyvec and \methodname, we construct their respective contextual feature hashes and for \bert we use the average of all token embeddings.
We report Spearman rank correlation based on cosine similarity as in \cite{reimers-etal-2016-task}.

\par
In the Pair Classification (PC) tasks, a model predicts whether two sentences are duplicates in a binary setting.
The metric reported for these PC tasks is the average precision based on cosine similarity.
These tasks are SprintDuplicateQuestions, TwitterSemEval2015, and TwitterURLCorpus.

\par
Finally, we include the Words in Context (WiC) \cite{wic} task for comparability with the accuracy scores reported in \cite{flyvec}.
Here, the goal is to disambiguate, in a binary sense, whether a specific word appearing in two sentences conveys the same meaning, given their context.
We closely follow the evaluation protocol explained in \cite{flyvec} when evaluating our models.

\par
For all tasks, we conduct a 5-fold cross-validation analysis.
Following \cite{flyvec}, we use only one fold to determine the hyperparameters for testing on the remaining four folds.
For the MTEB tasks, we optimize the hash length $k$. 
For WiC, we optimize the same hyperparameters as in \cite{flyvec}.

\subsection{Reinforcement Learning (RL) evaluation}
In practice, learning policies for RL environments require a significantly large number of environment interactions and training iterations.
We argue that \pflyvec and \methodname are computationally efficient, and thus, well-suited for this scenario.
RL involves simulation environments phrased as Markov Decision Processes (MDP) that involve a very large number of learning iterations.
For our evaluation, we use DDxGym \cite{ddxgym}, which is an RL environment focused on differential diagnosis -- the process of elimination that doctors employ to determine the diagnoses of patients.
This environment generates sequences of text describing the state of a patient.
Here, an agent learns a policy to choose \textit{diagnosis} or \textit{treatment} actions to cure a patient.
The \textit{diagnosis} actions uncover state information (e.g., symptoms) about the patient, while the \textit{treatment} actions treat them. 
Since not all the information about the state (symptoms, severity, and overall health condition) is present in each observation, this is a partially observable MDP (POMDP).
The environment is based on a knowledge graph of 111 diseases and an action space of 330 actions.
We compare the \textit{mean reward} and \textit{mean episode length} of all models in this environment. 
Both metrics show the level of success to diagnose and treat a simulated patient given a textual representation of their state. 

\par
All evaluated models use the auxiliary language modeling objective described in \cite{ddxgym}.
In the case of \pflyvec and \methodname, this objective is based on the energy in \cref{eq:E} and \cref{eq:E-complex}, respectively.
For the action and value function prediction, we use a projection layer on the hashes, ignoring gradient computation for $\vect{W}$.
This means that $\vect{W}$ is only fine-tuned by the auxiliary language modeling objective.
To compare with the results in \cite{ddxgym}, we use \texttt{IMPALA} \cite{espeholt2018impala} to optimize our policy for a maximum of 80M steps.
This algorithm is a state-of-the-art actor-critic method to parallelize the optimization of a policy across multiple workers (GPUs).

\subsection{Hyperparameters and implementation}
When compared, we use the same number of parameters for \flyvec and \methodname in all experiments.
Namely, 400 Kenyon cells ($K=400$) and their respective tokenizers: \numprint{20000}-word indices for OWC and \numprint{30522} for PubMed, i.e., $\nvoc=\numprint{20000}$ and $\nvoc=\numprint{30522}$, respectively.
We follow \flyvec to pretrain our models for 15 epochs with a learning rate of $4\times10^{-4}$, which is linearly annealed towards $0$.

\par
Our models are developed with \texttt{PyTorch} \cite{paszke2019pytorch}.
For multiprocess parallelization and RL, we use \texttt{Ray} \cite{ray} and \texttt{RLlib}\cite{liang2018rllib}.
For the WiC task, we conduct hyperparameter tuning using the \texttt{hyperopt}\cite{hyperopt} scheduler.
For training, we use \texttt{Adam} \cite{adam} as optimizer.
The sparse matrix multiplications in our model are mainly indexing operations, which we favor over the sparse\footnote{\url{https://pytorch.org/docs/stable/sparse.html}} abstractions in \texttt{PyTorch}, due to a much lower memory footprint. 

\par
For pretraining, we use eight A100 GPUs and a batch size of 0.8 \textit{million} samples.
In contrast, the sentence similarity evaluation on \flyvec and \methodname is run on 64 CPU cores to compute all hashes.
For \bert we use one V100 GPU and a batch size of 8 in the evaluations.
In the RL evaluation, for \pflyvec and \methodname we use four P100 GPUs and the same learning rate as in pretraining.
For \transmlm we report the results stated in \cite{ddxgym}.

\section{Results and Discussion}
In the following sections, we present the results of our experiments in both semantic similarity and Reinforcement Learning (RL) tasks.

\subsection{Semantic similarity}
\Cref{tab:mteb} presents our results for the semantic similarity tasks.
\begin{table*}[tb]
    \tiny
    \centering
    \caption{Five-fold cross-evaluation of Semantic Textual Similarity (STS) (top), Pair Classification (middle), and Words in Context (WiC) result for \bert, \flyvec, \methodname and \methodnametwo with the best hash length $k$ for each task.
    Except for STS12, our method outperforms \flyvec using the same number of parameters.
    \methodname also outperforms \bert in seven out of the 13 tasks.
    In terms of evaluation time, \flyvec, \methodname and \methodnametwo perform similarly and broadly outperform \bert even though the latter was evaluated on a GPU.
    The best score per task is highlighted in \textbf{bold}, and the best times are \underline{underlined}.
    The results with an asterisk* are taken from \cite{flyvec}.
    }
    \begin{tabular}{lcccc|cccc}
    \toprule
         & \multicolumn{4}{c}{Score mean $\pm$ std ($\uparrow$)} & \multicolumn{4}{c}{Evaluation time [s] mean $\pm$ std ($\downarrow$)} \\
         \midrule
         Task & \bert & \flyvec & \methodname & \methodnametwo &\bert & \flyvec & \methodname & \methodnametwo \\
        \midrule
        BIOSSES & \textbf{44.4$\pm$4.64} & 18.4$\pm$4.07 & 28.7$\pm$7.49 & 26.7$\pm$8.11 & \underline{0.28$\pm$0.01} & 0.49$\pm$0.01 & 0.45$\pm$0.02 & 0.76$\pm$0.11 \\
        SICK-R & 51.4$\pm$1.78 & 42.0$\pm$1.97 & \textbf{53.9$\pm$1.59} & 52.7$\pm$1.87 & 11.32$\pm$0.26 & \underline{3.43$\pm$0.11} & 3.68$\pm$0.35 & 4.25$\pm$0.35 \\
        STS12 & 33.8$\pm$7.04 & \textbf{46.3$\pm$7.49} & 45.5$\pm$7.33 & 42.4$\pm$7.10 & 4.27$\pm$0.19 & 1.59$\pm$0.18 & 1.67$\pm$0.08 & \underline{1.59$\pm$0.16} \\
        STS13 & 48.8$\pm$3.76 & 42.1$\pm$2.13 & 51.8$\pm$1.70 & \textbf{52.7$\pm$1.99} & 2.56$\pm$0.50 & 1.19$\pm$0.07 & 1.23$\pm$0.08 & \underline{1.13$\pm$0.09} \\
        STS14 & 44.3$\pm$3.52 & 42.1$\pm$2.61 & \textbf{53.6$\pm$2.25} & 51.9$\pm$2.68 & 6.69$\pm$1.24 & \underline{1.72$\pm$0.16} & 1.83$\pm$0.06 & 1.74$\pm$0.13 \\
        STS15 & 57.3$\pm$2.88 & 53.5$\pm$2.76 & \textbf{60.4$\pm$1.24} & 59.5$\pm$0.90 & 5.54$\pm$0.41 & \underline{1.57$\pm$0.10} & 1.73$\pm$0.07 & 1.69$\pm$0.09 \\
        STS16 & \textbf{60.2$\pm$6.21} & 45.5$\pm$6.30 & 49.7$\pm$4.53 & 47.1$\pm$4.35 & 2.26$\pm$0.17 & 1.08$\pm$0.08 & 1.10$\pm$0.09 & \underline{1.04$\pm$0.04} \\
        STS17 & 64.5$\pm$2.26 & 52.3$\pm$6.93 & \textbf{67.7$\pm$1.45} & 59.5$\pm$12.04 & \underline{0.23$\pm$0.01} & 0.74$\pm$0.06 & 0.63$\pm$0.07 & 0.95$\pm$0.07 \\
        STSBenchmark & \textbf{45.4$\pm$3.19} & 33.8$\pm$4.90 & 43.3$\pm$2.32 & 41.0$\pm$2.71 & 1.83$\pm$0.02 & \underline{1.14$\pm$0.07} & 1.27$\pm$0.10 & 1.42$\pm$0.45 \\
        \midrule
        SprintDuplicateQuestions & 31.6$\pm$0.90 & 40.4$\pm$0.70 & \textbf{56.2$\pm$1.02} & 55.0$\pm$1.09 & 11.98$\pm$0.18 & \underline{7.78$\pm$0.11} & 8.22$\pm$0.10 & 8.07$\pm$0.35 \\
        TwitterSemEval2015 & \textbf{51.3$\pm$1.57} & 38.4$\pm$1.74 & 51.2$\pm$2.66 & 50.8$\pm$2.00 & 9.90$\pm$0.13 & \underline{2.75$\pm$0.08} & 2.98$\pm$0.19 & 3.03$\pm$0.17 \\
        TwitterURLCorpus & 71.1$\pm$1.08 & 66.0$\pm$0.93 & \textbf{75.4$\pm$1.14} & 74.6$\pm$1.23 & 84.89$\pm$5.15 & \underline{7.24$\pm$0.84} & 8.00$\pm$0.54 & 7.64$\pm$0.23 \\

        \midrule
        WiC                      & \textbf{61.2$\pm$0.22}* & 57.7$\pm$0.27* & 57.08$\pm$1.10 & 58.16$\pm$1.59 & - & - & - & - \\
        \bottomrule
    \end{tabular}
    \label{tab:mteb}
\end{table*}

\paragraph{Task scores.}
The scores show that on seven out of 13 datasets, \methodname outperforms all other methods.
Except for STS12, \methodname consistently outperforms \flyvec by a large margin, signaling that for these tasks the added positional information is beneficial.
\bert outperforms on five out of 13 tasks. 
However, in two of these, TwitterSemEval2015 and STSBenchmark, \methodname is very close in second place.
\flyvec outperforms in STS12, where \methodname follows closely.
Similarly, \methodnametwo outperforms \flyvec and \bert in six tasks, however falls short of \methodname on all tasks except for STS13.
Nevertheless, these results show that \methodnametwo is an additional viable method to compute sentence hashes.

\par
It is remarkable that \methodname manages to bridge the performance gap and improve on \bert with only 14.56\% of the parameters, and that this is achieved solely by incorporating positional information.
Note that only unsupervised pretraining is involved and no post-hoc methods of model compression \cite{compression} are employed.
In practice, the latter is the most popular avenue to achieving significantly smaller models with comparable performance.

\paragraph{Task evaluation time.}
The evaluation times of \flyvec and \methodname on CPUs are significantly smaller than those of \bert on GPU.
BIOSSES and STS17 are the only tasks for which \bert shows a comparatively shorter time.
We argue that this is solely due to the small number of samples in these tasks. 
This effect is confirmed with datasets like TwitterURLCorpus, where the number of samples is significantly larger.
Here, the efficiency benefits of \flyvec and \methodname variants become evident (only ${\approx}7$ compared to ${\approx}85$ seconds of evaluation time).
We argue that the differences in efficiency observed in the evaluation time of \flyvec, \methodname, and \methodnametwo are mainly due to tokenization, hardware optimization, and implementation details.
Still, in terms of the hash computation both \methodname and \methodnametwo are mathematically more complex as can be inferred from the times of larger tasks, such as TwitterURLCorpus.

\par
This is reinforced when considering pretraining time.
\methodname takes approximately 15 minutes per epoch, which is significantly higher than the reported time for \flyvec on the OWC corpus -- eight minutes on less powerful hardware ($8\times\text{A100}$ vs. $3\times\text{V100}$) \cite{flyvec}.
This difference comes from our choice of implementation.
We favor an automatic differentiation framework, while \cite{flyvec} computes analytically a learning rule to fit the parameters.
Even in this setting, our training resources of ${\approx}4$ hours on $8\times\text{A100}$ are significantly inferior to the ${\approx}96$ hours of pretraining of \bert on 16 TPUs reported in \cite{bert}. 
When we pretrain \pflyvec for the RL evaluation with the PubMed corpus with our implementation (optimizing \cref{eq:E}), we report that \pflyvec and \methodname take roughly 12 and 15 minutes per epoch, respectively, highlighting the added complexity of our energy proposed in \cref{eq:E-complex}.

\begin{figure*}[tb]
    \centering
    \includegraphics[width=\linewidth]{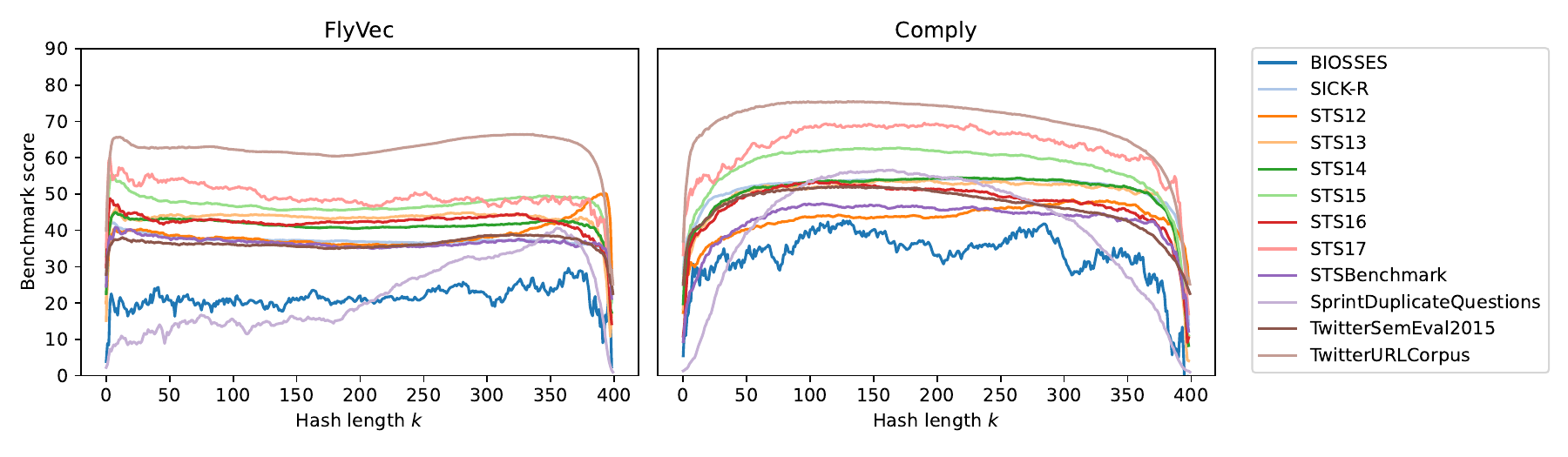}
    \caption{Test metric averaged over five folds with respect to the hash length $k$ on the STS and PC tasks.
    \methodname shows a better and more regular performance with respect to $k$, plateauing at around $k=150$, while \flyvec tends to perform better on either sparse (low $k$) or dense (high $k$) hashes for these tasks.}
    \label{fig:hash-lengths}
\end{figure*}

\begin{figure}[tb]
    \centering
    \includegraphics[width=0.45\linewidth]{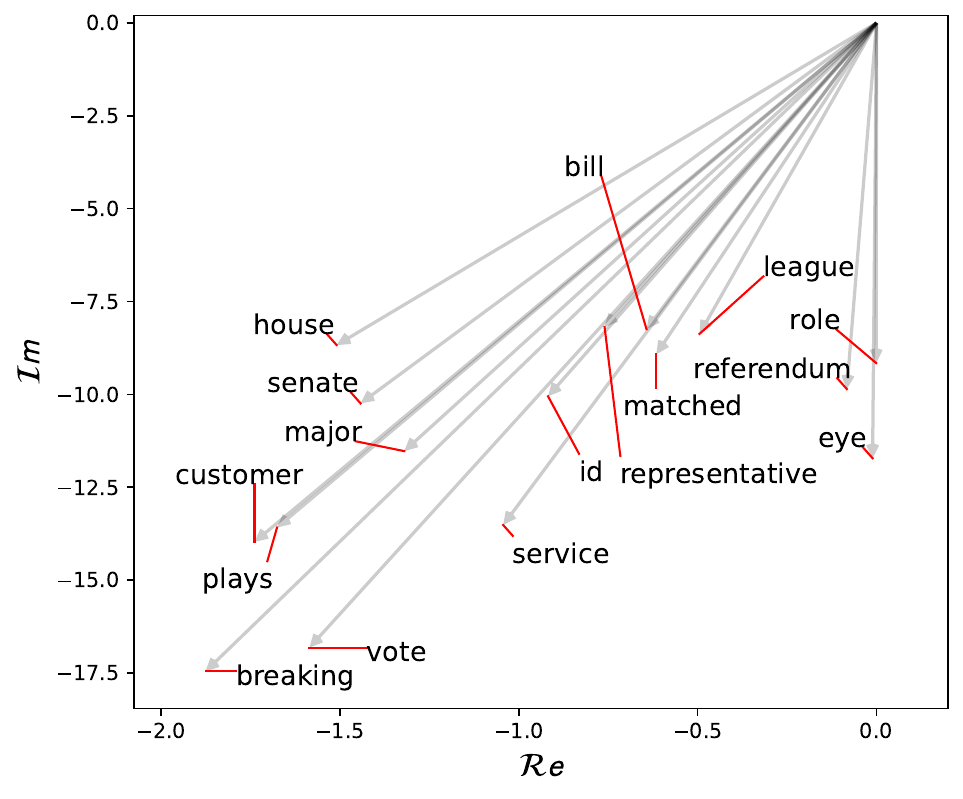}
    \caption{Highest activated neuron on the input sentence \enquote{\textit{Senate majority leader discussed the issue with the members of the committee}} as in \cite{flyvec}. The neuron shows signs of memorizing (overlapping) of short sequences like $[\texttt{house}, \texttt{representative}]$, $[\texttt{plays}, \texttt{major}, \texttt{role}]$, and $[\texttt{senate}, \texttt{bill}]$, among others. These are meaningful to the context of the input and likely to appear in the pretraining corpus.}
    \label{fig:neuron_ex}
\end{figure}

\paragraph{Hash length.}
Additionally, we survey the relationship between the mean test performance in the MTEB tasks and the chosen hash length $k$.
We present this in \cref{fig:hash-lengths} comparing \flyvec and \methodname.
Notably, the scores of \methodname are not only higher, but also more regular and concave, allowing for a hash length $k$ that is optimal with sparsity in the neighborhood of $150$ KCs.
In contrast, \flyvec tends to have an optimal $k$ either with very low or high values (very high and low sparsity).
Minimizing the energy of \methodname in \cref{eq:E-complex} can be interpreted as memorizing sequences.
Hence, we argue that the relationship between the scores and $k$ is due to these sequences being partially distributed across multiple neurons.
Although the energy optimized in \flyvec could accomplish the same (due to the sliding window, and as confirmed with the tasks that perform better with very dense hashes), we argue that presenting the additional positional information explicitly in $\C$ creates more distinct hashes.
We observe these partially distributed sequences in \methodname and present a qualitative example in \cref{fig:neuron_ex} of the most activated neuron conditioned on a specific sentence taken from \cite{flyvec}.
This neuron shows contextually relevant short sequences that can be extracted due to the relative position preserved in the phases of the words.
Examples of such sequences can be found in the caption.

\subsection{Reinforcement Learning (RL)}
\begin{figure}[tb]
    \centering\includegraphics[width=0.45\linewidth]{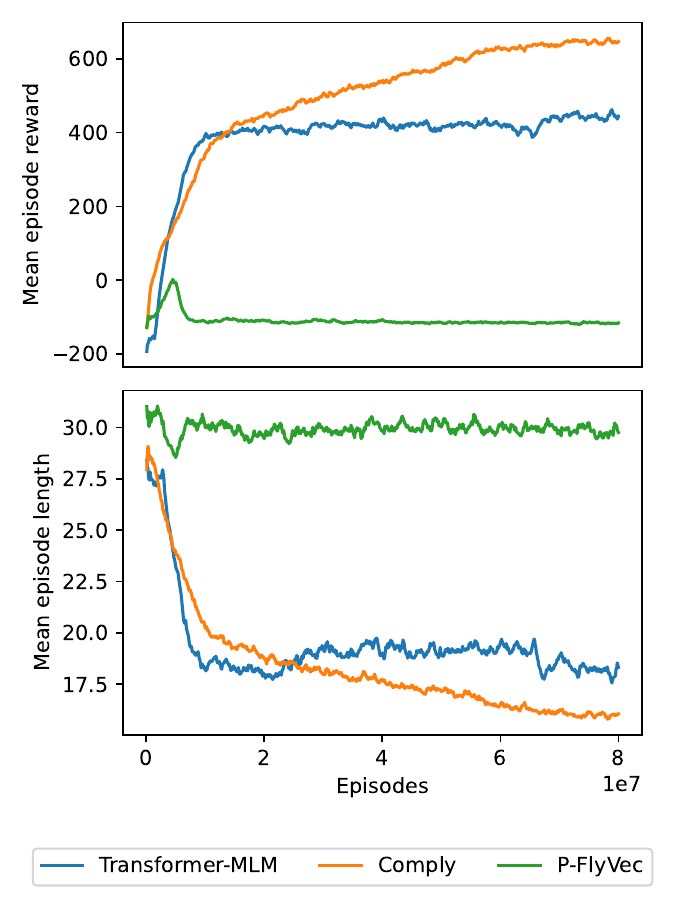}
    \caption{Mean episode reward (top, higher is better) and mean episode length (bottom, lower is better) in DDxGym. \methodname outperforms and continues to improve, while the Transformer baseline converges/peaks at around 15M episodes.}
    \label{fig:rl_results}
\end{figure}

In \cref{fig:rl_results}, we present the results of the evaluation on DDxGym \cite{ddxgym}.
Both mean episode reward and mean episode length estimate how well the models learn to solve this environment.
For the Transformer baseline both metrics are taken from \cite{ddxgym}.
\methodname outperforms the Transformer baseline roughly after 15M environment steps, and it continues to improve until the 80M mark.
The Transformer converges to a mean reward of 400 after roughly 10M steps.
Since the theoretical maximum for the reward in DDxGym is at about 1200, there is significant room for improvement.
The mean episode length shows a similar behavior -- here a lower value means that the episode ends successfully with the diagnosis both uncovered and treated. 
While the Transformer converges at around 18 steps per episode, \methodname improves until the experiment stops at 80M episodes with around 16 steps per episode.
Although \pflyvec initially reaches a reward of almost 0, it fails to improve further and shortly collapses in performance.
This further confirms that capturing positional information significantly improves the performance of \methodname, and makes the method more competitive to larger sequence-to-sequence models like Transformers.

\section{Conclusions}\label{sec:conclusion}
We present \methodname, an approach to improve learning of sequences by introducing positional information to a biologically-motivated model of the mushroom body of a fruit fly.
We achieve this by a novel loss that leverages representations in the complex field $\mathbb{C}$ and generalizes \flyvec \cite{flyvec} to learning sentences.
Our experiments show that \methodname improves sequence representations without adding more parameters.
In sentence similarity tasks we significantly bridge the gap between these biologically-inspired models and larger Transformer encoders like \bert. 
We show that the resource efficiency of our method makes it suitable for challenging settings like Reinforcement Learning (RL), where it outperforms a Transformer baseline.
Our analysis expands on the mechanism of the loss we introduce and the interpretability of the learned feature representations.

\section{Limitations and future work}\label{sec:future}
\paragraph{Tasks and modalities.}
We limit our experiments to pretraining our model on large general-purpose corpora.
Including the objective of sequence similarity as in sentence Transformers \cite{s_bert} would expand the comparison to such models.
Although our work presents results on the textual modality for comparability with previous work, we believe that our method is applicable to different modalities.
Thus a more extensive evaluation of, e.g., different RL environments or time-series tasks is needed.

\paragraph{Loss and model design.}
Word repetition is supported in our model in a limited way.
Namely, the phases of repeated words are averaged by the aggregation (see \cref{eq:E-complex}) during training.
Improvements in our method could stem from incorporating word repetition as a factor when learning sub-sequences across neurons as we show qualitatively in \cref{fig:neuron_ex}.
Furthermore, for the sake of comparability, we do not extend our evaluation to larger models (larger $K$). 
Scaling induces problems such as \textit{dead neurons}, so strategies like the GABA switch as proposed by \cite{sdm} or similar extensions could be needed.
Thus, future work would involve a study on the effect of scale on our approach.
Finally, the inherent benefit of adding positional information into a learned representation yields the potential for sequence generation.
We do not explore this in our work, but we see making the model generative as a future direction of research.

\begin{ack}
Our work is funded by the German Federal Ministry of Education and Research (BMBF) under the grant agreements 01|S23013C (More-with-Less), 01|S23015A (AI4SCM) and 16SV8857 (KIP-SDM). This work is also funded by the Deutsche Forschungsgemeinschaft (DFG, German Research Foundation) Project-ID 528483508 - FIP 12, as well as the European Union under the grant project 101079894 (COMFORT - Improving Urologic Cancer Care with Artificial Intelligence Solutions). Views and opinions expressed are however those of the author(s) only and do not necessarily reflect those of the European Union or European Health and Digital Executive Agency (HADEA). Neither the European Union nor the granting authority can be held responsible for them.
\end{ack}

\bibliography{main}{}
\bibliographystyle{apalike}

\end{document}